\documentclass[conference]{IEEEtran}
\usepackage{amsmath,amssymb,amsfonts}
\usepackage{algorithmic}
\usepackage{graphicx}
\usepackage{textcomp}
\usepackage{xcolor}
\def\BibTeX{{\rm B\kern-.05em{\sc i\kern-.025em b}\kern-.08em
    T\kern-.1667em\lower.7ex\hbox{E}\kern-.125emX}}

\begin{document}

\title{Geometric Perception based Efficient Text Recognition\\
\thanks{Identify applicable funding agency here. If none, delete this.}
}

\author{
\IEEEauthorblockN{P.N.Deelaka}
\IEEEauthorblockA{\textit{Computer Science and Engineering} \\
\textit{University of Moratuwa}\\
Colombo, Sri Lanka \\
180127u@uom.lk}
\and
\IEEEauthorblockN{D.R.Jayakody}
\IEEEauthorblockA{\textit{Computer Science and Engineering} \\
\textit{University of Moratuwa}\\
Colombo, Sri Lanka \\
180259b@uom.lk}
\and
\IEEEauthorblockN{D.Y.Silva}
\IEEEauthorblockA{\textit{Computer Science and Engineering} \\
\textit{University of Moratuwa}\\
Colombo, Sri Lanka \\
180118t@uom.lk}
}

\maketitle

\begin{abstract}
Every Scene Text Recognition (STR) task consists of text localization \& text recognition as the prominent sub-tasks. However, in real-world applications with fixed camera positions such as equipment monitor reading, image-based data entry, and printed document data extraction, the underlying data tends to be regular scene text. Hence, in these tasks, the use of generic, bulky models comes up with significant disadvantages compared to customized, efficient models in terms of model deployability, data privacy \& model reliability. Therefore, this paper introduces the underlying concepts, theory, implementation, and experiment results to develop models, which are highly specialized for the task itself, to achieve not only the SOTA performance but also to have minimal model weights, shorter inference time, and high model reliability. We introduce a novel deep learning architecture (GeoTRNet), trained to identify digits in a regular scene image, only using the geometrical features present, mimicking human perception over text recognition. The code is publicly available at https://github.com/ACRA-FL/GeoTRNet

\end{abstract}

\begin{IEEEkeywords}
Scene Text Recognition, federated learning, metric learning
\end{IEEEkeywords}

\begin{figure*}[!t]
  \centering
  \includegraphics[width=0.98\textwidth]{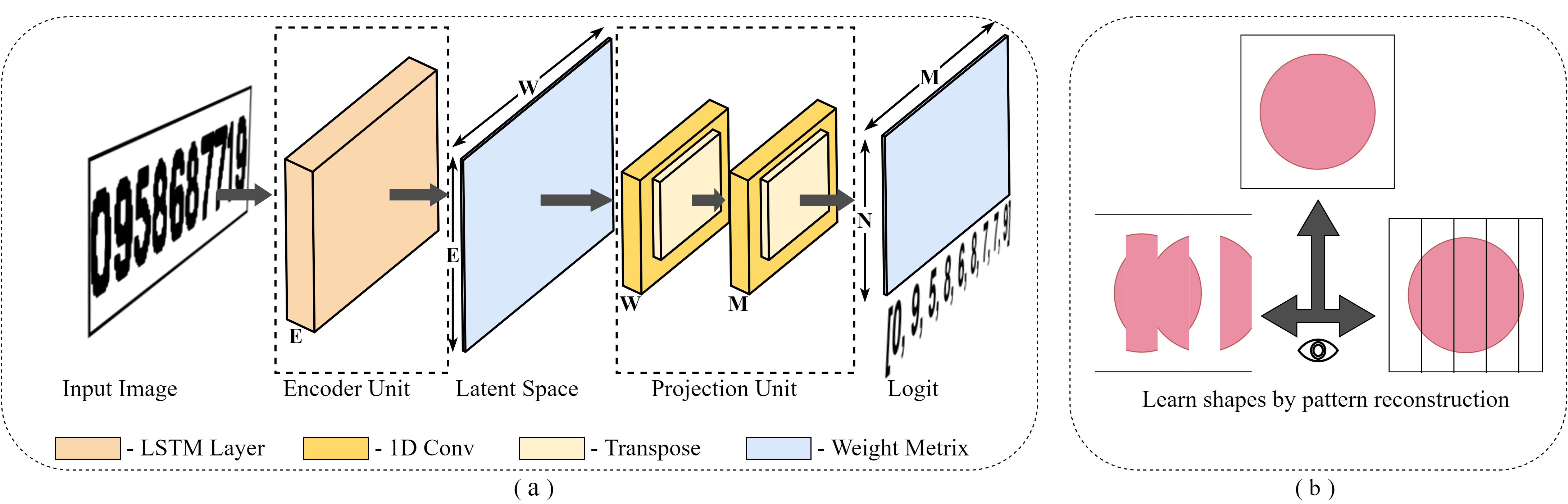}
  \caption{end-to-end model architecture composite of encoder unit and projection unit, and with real output heatmap.}
  \label{fig:model_archi}
\end{figure*} 

\section{Introduction} \label{intro}
The task of Optical Character Recognition (OCR) has been widely explored within the field of computer vision. Every character recognition task can be classified into two subtasks: character localization and character classification. With the introduction of the Region Proposal Network (RPN) \cite{p_rpn} and other advancements in object detection, the performance of the character spotting task has been steadily increasing. Also, with the attention mechanism \cite{p_attnOD} and autoencoder-based segmentation mechanisms \cite{p_detnet},the performance of the character recognition task improved.

However, this performance improvement was often accompanied by a significant increase in model size and computational complexity, despite the underlying task remaining the same. Specifically, State Of The Art (SOTA) models such as RF-Learner \cite{p_rfl} and  SPIN \cite{p_spin} have the parameter space on the scale of $10^6$. The usage of sizable models can be problematic in situations where real-time or near real-time operation is required. For example, deploying models on edge devices or training models in a federated learning fashion can benefit greatly from having a lightweight model architecture.

In this paper, we introduce an efficient and high-performance Deep Learning (DL) model architecture that is specifically designed to minimize the model parameter space. In addition to the very high inference speed, due to the significantly small model size, the model could be deployed on edge devices and even be trained on edge devices in a federated learning fashion. Since data privacy is a high priority and data drift is an unavoidable hazard in Machine Learning (ML), training an OCR model in a federated learning schema is profitable in both research and industry contexts. With the introduction of our architecture, the OCR model could achieve SOTA performance via federated training, as well.

Unlike other image recognition tasks, the only significant factor in the text/character recognition task is the geometry of the character. Factors such as color, depth, or hierarchical features have minimal influence on the output of the task. However, most models extract localized features from the RGB pixel space without explicitly designing for an efficient feature-extracting process. As a result, to achieve higher accuracy, the model size has to be increased. Consequently, the model training sample space has to be expanded to minimize the estimation error in empirical risk minimization. As a solution, the proposed model extracts geometric features from the grayscale image to identify the sequence of characters, without explicitly processing character locations.

This paper introduces a novel model architecture composed of a novel image feature encoding strategy and feature projection methods. Furthermore, the model is trained on images with a fixed number of characters, and thus outputs a fixed number of object labels, without specific training or predicting the object locations/ bounding boxes. Ultimately, this is targeting significantly smaller parameter spaces, and less computationally complex operations to tackle the real-time inference scenarios and optimal training in a federated learning setup. The model consists of two novel implementations: (1) a geometrical shape-based encoder, and (2) a feature localization unit for ground-truth label sequence prediction.

The model assumes the input image only contains a row of characters in the horizontally aligned format. The proposed model’s input format is a standard image format in the Scene Text Recognition (STR) field. However, any other text alignment such as curved or tilted can be used for model training with a preprocessing mechanism such as Thin Plate Spline transformation (TPS) \cite{p_tps}. Furthermore, the proposed model can be used in an end-to-end text recognition task, simply by adding a bounding box detection model such as SegLink \cite{p_seglink} or BoundaryE2E \cite{p_boundary}  before the recognition model.

For the first novel feature, the paper introduces a very lightweight encoder unit that encodes the shape of each character found in the image. In the real world, we could safely say that someone learned a shape, by examining how someone can reorganize a striped image of a shape, back together [\ref{fig:model_archi}(b)]. As a deep learning counterpart, we propose LSTM-based encoders trained to learn the aforementioned principal idea. Furthermore, the following layer configurations and the objective function are defined to ensure the encoder learns as expected.

The second novelty is that the encoded feature space is mapped into a logit layer by a unique spatial attention mechanism. This submodel consists of several unique innovations to achieve SOTA performance. As unveiling those new implementations from input to end, the model is composed of an interpretable feature region to target the label mapping block. Not only is this block mathematically stable to prevent the model from collapsing but it’s also very lightweight compared to the general attention \cite{p_attnOD} mechanism. On the other hand, the paper introduces a fully-convolutional label prediction mechanism, loosely resembling the auto-encoder architecture \cite{p_autoe}. Furthermore, the proposed architecture reveals a superior logit layer format to traditional multi-class multi-label classification \cite{p_multil} logit format in several aspects. 

In addition to the novel model architecture, the paper also introduces a sophisticated synthetic text sample generation python library called digitgen. digitgen library builds as a solution to several drawbacks found in well-known datasets like SynthText \cite{d_sjsynth} and MJSynth \cite{d_mjsynth}. The implemented image synthesizing algorithm enables the generation of a dataset of any size, irrespective of the memory size of the system. Furthermore, it is able to generate a COCO-structured annotation file \cite{d_coco} to rebuild the data loaders under any deep learning framework. Additionally, the characters in the images are generated completely pseudo-randomly, unlike in MJSynth or SynthText. This random character definition is proposed to mitigate the propensity added by the lexicon-based scene text generation. Despite synthetic generation, the proposed library provides a variety of augmentations such as adding random noise, adding shadow patches, adding light bursts, changing font family, font size, etc.

To summarize, the paper introduces a novel model architecture to address the task of regular scene text recognition (GeoTRNet), and a new synthetic regular scene text image generating python library (digitgen). With the aforementioned introduction, the paper empirically proves the model’s SOTA performance under near real-world image datasets. Additionally, the experimental results under federated learning with low client sample sizes, serve as evidence for the model’s capability in estimation error minimization. Furthermore, the conducted experiments on metric learning for the signature verification task show the wide usability of the proposed architecture.

\section{Related Works} \label{rel}
The field of scene text recognition has gained a lot of progress throughout the years and many solutions model the task with a sequence prediction objective function.
Graves et. al. \cite{p_hwrecog} have approached this task by handcrafted geometrical features such as mean gray value and gradient of pixel intensity. Su and Lu \cite{p_hogstr} achieve the feature extraction using HOG features obtained from the cropped images of the words. Most later works have turned towards deep learning to replace hand-crafted feature extraction.
Shi et. al. \cite{p_crnn} were the first to utilize a combination of CNNs and RNNs to tackle the task of text recognition (CRNN). The CNN was used to extract a set of feature maps from the input image. The feature map was then broken down into one-pixel-wide columns of pixels and the $i^th$ pixel columns of each feature map were concatenated to form the feature vectors which were then fed into an RNN. They further improved on this by swapping the RNN for a Bidirectional LSTM which showed better performance. This setup proved to be highly compact and efficient due to the removal of fully connected layers.
Shi et. al. \cite{p_crnn} use a CTC layer \cite{p_ctc} to define the conditional probability of the sequence of labels. However, the computationally suboptimal nature of the CTC loss makes the model training highly volatile and time-consuming. To address this issue, Xie et. al. \cite{p_ace} introduced the Aggregation Cross-Entropy loss which was highly efficient.
In \cite{p_tpami}, Baek et. al. introduce a four-stage STR framework that most STR models fit into. Of these, the first stage is defined to be the irregular/curved text image to regular text image transformation. In the case of irregular text, the transformation stage is a significant necessity to ease the processing for subsequent stages. However, in this paper, we will build on tasks where the model receives regular text as input.
The task of transformation is handled by different models using various methods. Shi et. al. \cite{p_seglink} address the case of oriented text detection using the SegLink model which detects individual characters, their orientations, and the links between characters of the same word. BoundaryE2E \cite{p_boundary} detects the boundary points around the text and uses them to warp the image to horizontal text. A similar mechanism is used by Qiao et. al. \cite{p_tperceptron}, where the text boundaries and corners are detected using semantic segmentation, prior to the boundary point detection. Liu et. al. \cite{p_abcnet} designed ABCNet to predict Bezier curves that would fit the text on the image. This was followed by the BezierAlign module which aligned the text as required.
Zhang et. al. \cite{p_spin} achieve both chromatic and geometric rectification using the SPIN network. As noted in \cite{p_spin}, this network can work in complement with other transformation models or with a standalone recognition model such as the one proposed in this paper.

\begin{table*}[!ht]
 \centering
%   \begin{center}
\begin{tabular}{ |p{5cm}|p{1.5cm}|p{1cm}|p{1cm}|p{1cm}|p{1.5cm}|p{1.5cm}| } 
\hline
\textbf{Experiment Name} & \textbf{Accuracy(\%)} &\textbf{ mAP(\%)} &\textbf{ mDP(\%)} & \textbf{time(ms)} & \textbf{Input size} & \textbf{\# of Parameters} \\
\hline
(a)digitgen [no augmentation] & 99.91 & 99.94 & 99.95 & 0.502 & 224x28 & 31,090 \\
\hline
(b)digitgen [random inter-digit spacing] & 99.03 & 98.79 & 98.76 & 0.461 & 224x28 & 31,090 \\
\hline
(c)digitgen [space character] & 98.95 & 98.77 & 98.77 & 0.497 & 224x28 & 16,323 \\
\hline
(d)digitgen [dynamic background intensity] & 99.93 & 99.91 & 99.92 & 0.513 & 224x28 & 31,090 \\
\hline
(e)digitgen [dynamic text width] & 95.55 & 95.31 & 95.37 & 0.574 & 224x28 & 31,090 \\
\hline
\hline 
SOCR Dataset & 97.70 & 96.70 & 95.18 & 0.840 & 244x48 & 19,356 \\
\hline
SOCR with random spacing & 96.30 & 95.70 & 94.28 & 0.769 & 244x48 & 19,453 \\
\hline
MNIST & 97.65 & 97.70 & 97.28 & 0.547 & 244x48 & 11,546 \\
\hline
MNIST with random spacing & 91.51 & 91.23 & 91.10 & 0.347 & 244x48 & 16,546 \\
\hline
\end{tabular}
% \end{center}
\caption{Experiment Result Table}
\label{table:1}
\end{table*}

\section{Method} \label{method}
\subsection{Model architecture}
The architecture mainly consists of two high-level blocks; the image feature encoder and the object projection units. The image encoder unit operates as the backbone model (similar to Faster R-CNN \cite{p_frcnn} encoder unit) and outputs a latent space that is directly fed into the projection layer. Instead of having two headers to predict possible bounding boxes and the class labels for those bounding boxes, here the proposed architecture only consists of an object class classification header. The underlying concept here is that since, in general, the computer-generated text will be horizontally aligned or could be directly transformed into a horizontally aligned format like in E2EBoundary \cite{p_boundary}, models just need to predict all the text present in an orderly fashion; from left to right.  

\subsubsection{Image encoder}
The encoder unit in the model targets the processing and extracting of spatial features, similar to how time-series models extract temporal features from data. In implementation, experiments with a Bi-LSTM, LSTM combination, and a temporal convolution network layer (CTN) \cite{p_tcn1}, \cite{p_tcn2} show the expected behavior of training with significant performance. On the other hand, the resulting encoding unit features are much denser than features generated by the usual CNN-based encoder.

As we defined in the introduction [\ref{intro}], the core concept here is to extract only the essential features for detecting the object class. In the proposed input configuration, model interpretability and reliability will substantially improve. Furthermore, the resulting model size will be significantly lower, and with that, inference speed will improve.
The underpinning of the architecture is to detect the text object, considering only the character shape/geometry. Also, geometry is the most important feature when a human identifies a character, unlike in any other task such as face recognition \cite{p_frec} or bird classification \cite{p_bclass}. First, the model only accepts the gray-scaled images to avoid the model having any feature other than the intensity value that is essential to detect the shape. Second, the encoder units consist of spatial feature encoder units only, unlike using convolution layers to process local features.

The LSTM layer processes the image, a single column of pixels at a time to understand the pixel pattern to extract features related to character shapes/geometry present in the image. The core idea of using the LSTM or TCN layer here is to identify the geometrical shapes \& extract related features without defining any manual geometrical feature extraction template. In analogy, the LSTM layer learns geometry by trying to organize a set of stripped parts of geometry [such as a circle] back together ~\ref{fig:model_archi}.

In the LSTM layer-based encoder configuration, the input is fed to the Bi-LSTM layer followed by a single LSTM layer, and the output is considered as the latent feature space. The main reason the encoder starts with Bi-LSTM is that empirically, Unidirectional LSTM layers show low confidence when identifying similar shapes such as character 3 from characters 8 \& 9. 
Theoretically, these confusing features are generated because the LSTM layer gets identical input columns up to a certain point. However, with the Bi-LSTM, theoretically, this problem is solved \cite{p_bilstm} because there could only be a single class that shows such a pattern in both directions (left-to-right \& right-to-left).

Also, the paper introduces TCN as the alternative layer to the LSTM-based layer configuration. Specifically, the TCN layer used in the implementation is an acausal process, unlike the generally used TCN layer which uses causal processing \cite{p_tcn2}. Acausal TCN essentially works similarly to the Bi-LSTM layer, with hierarchical feature processing \cite{p_tcn2} as compensation for not having a gated weighting system as in LSTM. Furthermore, empirical results with the TCN layer prove the theoretical similarities between LSTM configuration and the TCN layer in the defined context.

\subsubsection{Object projection unit}
There are two main tasks handled by the introduced projection unit. First, locate the regions of feature space related to each character in the image. Secondly, predict the class related to each character in the same order as the characters in the input image.

The first task is achieved by implementing a 1-dimensional convolution (1D Conv) \cite{p_1dconv} with an output channel for each target class label [Figure~\ref{fig:model_archi}]. The concept is to define each class probability/logits based on the features from the latent space. In this way, the encoder unit will be only responsible for the feature generation process, and the projection unit is relieved from the feature extraction task. The loosely coupled nature of the proposed architecture will be essential for having solid, well-interpretable backpropagation steps in the train time. 

Also, the model achieves a second task by another 1D Conv layer with a channel per character in the input image. The intuition here is to convolve the class probabilities based on the relative locations. Furthermore, the model applies the matrix transpose operation before and after the second 1D Conv process. The order of the two 1D Conv layers is significant because it's crucial to have probabilities before they are assigned to resulting character positions. This is because, with the proposed order, their information loss will be much lower than in other ways.

In addition to the projection unit, here the paper introduces a novel logits layer structure because existing logits layer structures such as multiclass or multi-label multi-class do not comply with the requirements. With the multilabel logits layer structure \cite{p_multil}, the model cannot separately interpret two objects under the same class. It is only able to predict each class appearance in the input image without any order. Hence, the paper introduces a 2D logits layer with rows with softmax activation \cite{p_softmax}, intuitively where each row represents class probabilities for each character in the image. 

As a whole, the projection unit works as a 2D attention unit, where each 1D Conv layer intuitively performs as a 1D attention unit \cite{p_1dattn}. The second 1D Conv layer’s convolution operation works as the spatially weighted average of class probabilities for characters, while the initial 1D Conv layer works as an attention unit to derive class probabilities from latent space feature columns. In addition to the novel attention mechanism in the projection unit, with the proposed layer configuration, the output size of the unit will be independent of the latent space size or input image size. 

\begin{figure*}[!t]
  \centering
  \includegraphics[width=0.98\textwidth]{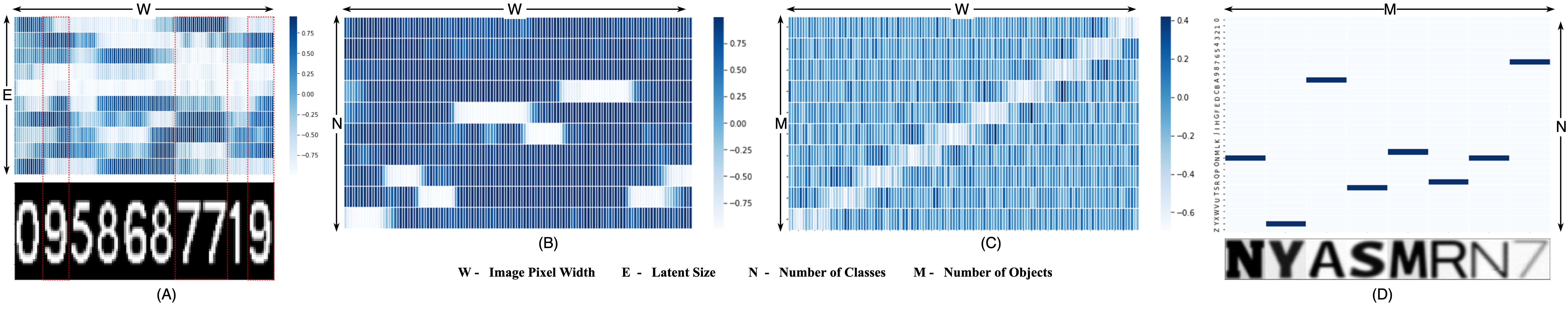}
  \caption{(a) GeoTRNet latent space output for the sample input sticker. (b) Spatial weights in the trained projection unit's second layer. (c) Class probability distribution outputs by the projection units layer. (d) Test Standard-OCR image with model predicted logits heatmap. }
  \label{fig:model_inter}
\end{figure*} 

\subsection{Model objective \& training}
The model trains under the objective function of categorical cross-entropy  (Equation \ref{loss_equ}). Where N denotes the number of label classes and M denotes the maximum number of text objects in a single input image. The only update introduced to the categorical cross-entropy is the sum over each. Even though the model emphasizes the object order, we do not define an explicit objective to evaluate order. This is because to reduce the loss, the model has to be able to predict the correct class at the correct position. Therefore overall, the model will predict the correct class per each object/digit present only by considering the geometrical properties of the localized region.

\begin{equation} \label{loss_equ}
    Loss = - \frac{1}{N} \sum_{m=1}^{M} \sum_{n=1}^{N} y_n.\log(\hat{y_n})
\end{equation}

An extensive hyper-parameter tuning process has shown the model optimally with the Adam\cite{p_adam} algorithm as the model optimizer. However, theoretically, GeoTRNet architecture does not support common model generalization mechanisms such as batch normalization \cite{p_bsn}, Dropout \cite{p_dropout} or stochastic depth\cite{p_sdepth} because of the restricted layer configuration. Thus, the model trains with sharpness aware minimization (SAM) \cite{p_sam} to improve the model generalization ability by leveraging the geometry of the loss landscape.

\subsection{Model inference}
The Model will only be able to output a fixed number of objects per image under any of the already seen classes in the training phase. Especially, if the model needs to identify space as a separate character, the training dataset has to be annotated with relevant spaces under a unique class. If not, the model will ignore the space, because the model passively uses spaces in the geometry detection process and feature localization operation, especially in the last 1D Conv layer.

\section{Experiment}
\subsection{GeoTRNet on Synthetic OCR digits dataset (digitgen)}

The model experiments are conducted on the cropped regular scene text image bounded to a predefined maximum object limit, hereby stated as stickers. The synthetic dataset MJSynth \cite{d_mjsynth} and the generators TextRecognitionDataGenerator \cite{d_text_rec_gen} and SynthText \cite{d_sjsynth} are used in the text recognition domain. Even though data synthesizing tools like MJSynth \cite{d_mjsynth} and TextRecognitionDataGenerator \cite{d_text_rec_gen} generate advanced scene text images, they are being generated based on lexicons and dynamic sticker object count, making them less suitable for the experimental usages. On the other hand, generators such as SynthText make multiple texts in a single image and random placement makes them incompatible for GeoTRNet standalone training. Due to these reasons, a proper comparison against existing architectures, which address the case of irregular scene text, is unattainable.

As a workaround to the previously mentioned incompatibilities, the paper introduces a python library, ‘digitgen’, for the generation of stickers using characters. The library has the ability to add specific augmentations to the sticker depending on user specifications. Due to the wide variety of augmentations that are enabled by digitgen, we are able to replicate the intricacies observed in actual regular scene text images.

Every experiment based on the ‘digitgen’ trains on a 10K sticker dataset while evaluating on an additional 2K sticker dataset. Also, all the model performance results are recorded on a separate 10K sticker test dataset on the best model from the training process, where each sample sticker is bound to 8 objects and resized to 224x28.

\begin{figure}[!t]
  \centering
  \includegraphics[width=0.98\linewidth]{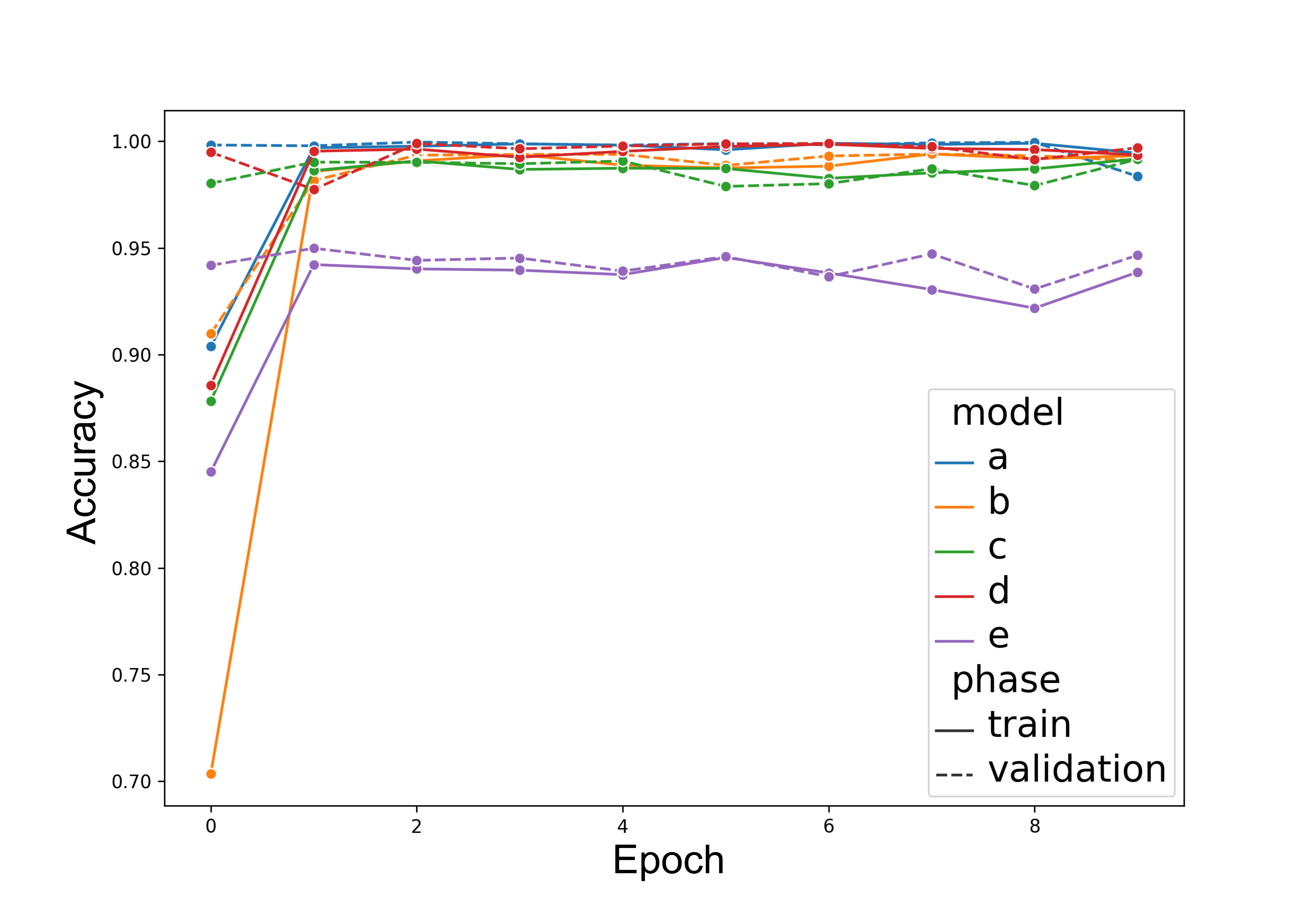}
  \caption{digitgen based GeoTRNet model training accuracy dynamics.}
  \label{fig:model_acc}
\end{figure}

According to the geometrical deep learning context \cite{p_geodl}, here we experiment to test whether GeoTRNet bounds the targeted functional space to minimize the approximation error while reducing the estimation error, even with the smaller sample space. The approximation error reduction is evident by the insignificant level of the model overfitting to any dataset we experimented with, even in federated learning settings [Figure~\ref{fig:model_acc}]. Furthermore, from the observed SOTA performance, it is safe to say that the trained model has enough variance, despite focusing on the estimation error as well.

Further, in the following experiments, major changes were applied to the input dataset format instead of the model architecture, to evaluate the model architecture’s performance in various real-time scenarios. A GeoTRNet model, with the image encoding done by a bidirectional LSTM layer with 48 cells followed by an LSTM layer with 48 cells, which hereby is defined as the base model, was trained on 10 epochs and tested on stickers generated by digitgen. For reference, the base model was trained without augmentation, achieving an accuracy of 99.91\% with 99.94% mAP and 99.95% mDP. An average inference time of 0.502 ms was recorded [Table \ref{table:1} (a)].

\subsubsection{Test on inter-digit space resilience}

The space between different characters can often vary based on fonts and other artistic effects applied to the text. In this experiment, we scale the inter-digit spaces within each sticker to random widths to mimic the real-world data which has characters spaced in an orderly fashion. The base model was trained and tested showing a significant performance of 99.03% accuracy with 98.79% mAP and 98.76% mDP. An average inference time of  0.461 ms was recorded  [Table \ref{table:1} (b)].

\subsubsection{Test on space as a character resilience}

To model the white space characters in real-world text scenarios, each sticker was generated with the inclusion of spaces/gaps between characters to which, a GeoTRNet model with the image encoding by a bidirectional LSTM layer with 48 cells followed by an LSTM layer with 48 cells was trained and tested, showing a significant performance of 98.95\% accuracy with 98.77\% mAP and 98.77\% mDP. An average inference time of  0.497 ms was recorded  with 16,323 parameters [Table \ref{table:1} (c)].

\subsubsection{Test on digit width resilience}

The case of varying digit widths, too, can be seen in real-world text due to how different fonts render characters. To model character size distortions the experiment was conducted where the digit width was changed at random. The base model was trained and tested achieving a significant performance of 95.55\% accuracy with 95.31% mAP and 95.37% mDP. An average inference time of  0.574 ms was recorded [Table \ref{table:1} (d)].

\subsubsection{Test on background intensity resilience}

The intensity of the background relative to the characters varies vastly based on the targeted visual look or even lighting conditions. To mimic the background intensity changes the background and digit intensities of each image were randomly altered. The base model was trained and tested showing a significant performance of 99.93\% accuracy with 99.91% mAP and 99.92% mDP. An average inference time of  0.513 ms was recorded[Table \ref{table:1} (e)].

Empirically GeoTRNet has proven its capability of handling the tested data format scenarios with at least 10 times the speed of real-time performance \cite{p_realtime} and with just a 31K parameter model.

\subsection{GeoTRNet on StandardOCR Sticker dataset}
Synthetic datasets are composed of a set of vector graphics but real-world datasets are composed of a set of raster images of vector graphics. Because of this fundamental difference, we implement an algorithm to build a near real-world, annotated dataset from the famous Standard OCR dataset (SOCR) \cite{d_stdOCR}. The generated dataset consists of a diverse set of stickers with various font types, font sizes, text styles, background intensities, and font color levels. A 12K image training set and a 2K image validation set were created using the SOCR training set. The extra 2K image test set created by the SOCR test image patch set was used for model evaluation purposes.

The trained model with ~19K parameters recorded 95.37\% accuracy values in the test datasets with 97.18\% mDP over ten digits. Compared to general text detection models such as CRNN \cite{p_crnn}, the trained model is 100X smaller in parameter space, even though the model archives SOTA performance. This performance proves the effectiveness of the underlying model architecture \& concept mentioned in the method section, in practice. Furthermore, as shown in Table[\ref{table:1}], the trained model response time for a single image detection (<0.8ms), is 10 times faster than real-time performance (6-20ms) \cite{p_realtime}.

\subsection{GeoTRNet on MNIST Sticker dataset}
In addition to testing the model on the targeted image distribution, here we experiment on the model with, geometry-wise, a much more complex dataset. We developed an algorithm to generate a sticker dataset from the well-known MNIST \cite{d_mnist} dataset, where each sticker corresponds to a handwritten list of numbers.  In the experiment, we generated 12K training images and 2K validation images from the train character set \& generated another 2K images for the test set using the MNIST test dataset. For each image, 8 random letter patches, each of size 28 by 28, were combined and resized to form a 244 by 48 sticker.

After a hyper-parameter tuning pipeline, with the optimal parameters, the model shows 97.65\% accuracy with 97.65\% mAP and 97.70\% mDP. 0.747 ms average inference time was recorded in the testing phase and where the model only contains ~11K parameters. Empirically, the model has proven its capability to process dynamic and inconsistent geometrical patterns like hand-written letters very well. 

\subsection{GeoTRNet Interpretation }
Since the paper introduces a novel architecture based on the entirely new image data processing and projection system, we follow an extensive model interpretability testing schema. We visualize the weights of the 1D Conv layers and their outputs for a sample input, interpret the results, and compare them with the expected results.

First, we visualize the resulting latent space for the input image as shown in Figure [Figure~\ref{fig:model_inter} (a)], where the latent vector size is set to 10 for easy visualization. The figure represents the spatial consistency of the feature by the continuous value patterns over the rows in the heatmap. Therefore, we could safely assume model encoders generate features primarily based on the character shapes. Furthermore, the Figure [Figure~\ref{fig:model_inter}(a)] illustrates the identicality of the features for the same target over different locations, especially for letter 9 duplicates. 

Secondly, the output feature matrix from the first 1D Conv layer [Figure~\ref{fig:model_inter}(b)] completely adheres to the expected behavior as stated in the method sections. The figure clearly shows the well-biased class probabilities over the span of feature columns. Also, the 1D Conv layer weight heatmap shows significantly similar weights over neighbors and applauds the idea of giving the same weights to adjacent neighbor features as processing feature columns.

Thirdly, the second 1D Conv weight space  [Figure~\ref{fig:model_inter}(c)] illustrates how the model assign / weighs based on spatial location. With the empirical evidence, it is safe to say the whole projection unit figuratively works as a 2D attention unit. Figure~\ref{fig:model_inter}(d) further emphasizes the location and class-based accuracy of the final predictions. Furthermore, here the input image used is a sample from the test set and the model hasn't seen the image in the training phase.

Finally, the model pruning result illustrates how dense the features are in the LSTM encoder unit. Here the model pruning is implemented using the Keras model pruning sub-module’s unstructured pruning process. Considering the overall result from the local model interpretability experiment, we could safely suggest that the model has high explainability compared to most other scene text recognition models \cite{p_mrpune}.

\subsection{Signature Verification task}
In addition to the field of STR, the model was experimented on the metric learning field to evaluate the model performance on the few shot settings. We experiment with the proposed architecture on the signature verification task on the CEDAR Signature dataset \cite{d_cedar}. The task is ideal for the model mainly because of the theoretical importance of spatial geometric feature recognition and naturally grayscale image input. 

The model training process follows the technique introduced by Elie et. la \cite{p_metl}. The only update to the proposed architecture in the metric learning setting is to add an embedding lookup layer to the model head. The model was trained on 884 images, with 18 samples per each of 55 signatures. Validation was performed with 436 images with 6 samples per each of the 55 classes of the training phase. The model trains under the cosine distance-based MultiSimilarity loss \cite{p_metl} with the Adam \cite{p_adam} optimizer. 

After the nearest-neighborhood-based distance calibration \cite{p_metl}, the model achieves an almost perfect F1-score and an accuracy score for the training dataset. For the test dataset, the model achieves an 86.67\% F1-score with 74.50\% accuracy. 
% As the figure [Figure .] shows, the model performs well in signature detection in practice with the unseen test data, as well.

\subsection{Adversarial Attack}
So far, from experiment results, the model showed impressive performance with a 100X smaller parameter space. We also discussed the model’s empirical performance on other tasks such as metric learning and federated learning. However, as stated in the Method \ref{method}, with the proposed architecture, the model’s reliability and trustworthiness improve significantly. 
We test the model’s resilience to the adversarial attack on the Fast Gradient Signed Method (FGSM) attack \cite{p_fgsm} with the model trained on the MNIST sticker dataset. In the paper \cite{p_fgsm}, Goodfellow et. al experimented with the FGSM method on a CNN-based classification model trained on the MNIST dataset where they showed that the model misclassification rate was recorded at 89.4\% with an epsilon of 0.25 and 87.15\% with an epsilon of 0.1. 
In our experiment, the model achieved 96.55\% accuracy on the test dataset. We recorded only a 7.90\% accuracy drop, even with epsilon set to 0.3. With the epsilon at 0.1, the accuracy drops experienced were less than 1\% on the adversarial example dataset built directly based on the scored test dataset. These values empirically prove the importance of restricting input space to a 2D intensity level and the reliability of the proposed architecture very well.

\section{Future Works}
The GeoRTNet model, even in its current form, opens a large number of doors with solutions in the trending deep learning topics such as privacy-aware model training, high-speed inference, and open-world object detection. Specifically, the simplistic and very-lightweight nature of the model lays the ground for optimal data distributed training in tasks such as credit-card security number reading and barcode number panel recognition. On the other hand, microsecond scale inference time comes in handy with tasks like highway vehicle number plate detection. Furthermore, the highly reliable nature and the well-explainable nature make the model suitable to operate on sensitive datasets.

However, the current model is restricted to only the digit recognition step on regular text scenes. The introduced model could be used to improve any multi-step or end-to-end STR model, just by replacing the model recognition header with the proposed model. This upgrade is guaranteed to provide impressive performance with drastic inference time reduction. Stated model upgrades can be done with minimal intervention in STR models such as BoundaryE2E \cite{p_boundary} and SegLink\cite{p_seglink}.

GeoTRNet merged with text localization and a pre-processing model \cite{p_boundary} could lead to significant improvements in the model accuracy, model reliability, and inference speed, because the highly reliable nature of the GeoTRNet, and the significant model weight reduction, even in the recognition header, leads to the stated claims respectively. As stated in the introduction, with transformations like SPT, the updated STR model supports irregular text images as well.

As empirically proven, the GeoTRNet model could also achieve outstanding performance in few-shot learning tasks such as federated learning and metric learning. The proposed model inherently could be used in tasks like signature verification, and hand-written letter quality/shape evaluation tasks, especially under any task that humans perceive solely from the object shape, with a very simple change to the core model, as shown in experiments.

\section{Conclusion}
In this paper, we introduced GeoTRNet, a novel model architecture for geometry-based text recognition. GeoTRNet is capable of recognizing a set of characters with SOTA accuracy within a fraction of a millisecond. This performance was maintained on text recognition tasks under standard and synthetic datasets with a model of only 1\% of the parameter size. Furthermore, the model was proven to perform well, even under the few-shot learning schemes such as federated learning and metrics learning. Also, GeoTRNet comes with high model explainability and very high resilience to adversarial attacks. Hopefully, the proposed architecture can open new doors in various other related tasks in the computer vision field. We are certain that the core concepts we introduced here could be used to build smart models in all areas of deep learning. 

\section{Acknowledgment}

We thank S.M.Chandanayake and L.T.N. Wickremasinghe for their support on this project. 

% \pagebreak
% \newpage
% \printbibliography %Prints bibliography 
\bibliographystyle{IEEEtran}
\bibliography{references}

\begin{thebibliography}{10}
\providecommand{\url}[1]{#1}
\csname url@samestyle\endcsname
\providecommand{\newblock}{\relax}
\providecommand{\bibinfo}[2]{#2}
\providecommand{\BIBentrySTDinterwordspacing}{\spaceskip=0pt\relax}
\providecommand{\BIBentryALTinterwordstretchfactor}{4}
\providecommand{\BIBentryALTinterwordspacing}{\spaceskip=\fontdimen2\font plus
\BIBentryALTinterwordstretchfactor\fontdimen3\font minus
  \fontdimen4\font\relax}
\providecommand{\BIBforeignlanguage}[2]{{%
\expandafter\ifx\csname l@#1\endcsname\relax
\typeout{** WARNING: IEEEtran.bst: No hyphenation pattern has been}%
\typeout{** loaded for the language `#1'. Using the pattern for}%
\typeout{** the default language instead.}%
\else
\language=\csname l@#1\endcsname
\fi
#2}}
\providecommand{\BIBdecl}{\relax}
\BIBdecl

\bibitem{p_rpn}
B.~Li, J.~Yan, W.~Wu, Z.~Zhu, and X.~Hu, ``High performance visual tracking
  with siamese region proposal network,'' in \emph{Proceedings of the IEEE
  conference on computer vision and pattern recognition}, 2018, pp. 8971--8980.

\bibitem{p_attnOD}
J.-S. Lim, M.~Astrid, H.-J. Yoon, and S.-I. Lee, ``Small object detection using
  context and attention,'' in \emph{2021 International Conference on Artificial
  Intelligence in Information and Communication (ICAIIC)}.\hskip 1em plus 0.5em
  minus 0.4em\relax IEEE, 2021, pp. 181--186.

\bibitem{p_detnet}
Z.~Li, C.~Peng, G.~Yu, X.~Zhang, Y.~Deng, and J.~Sun, ``Detnet: Design backbone
  for object detection,'' in \emph{Proceedings of the European Conference on
  Computer Vision (ECCV)}, September 2018.

\bibitem{p_rfl}
H.~Jiang, Y.~Xu, Z.~Cheng, S.~Pu, Y.~Niu, W.~Ren, F.~Wu, and W.~Tan,
  ``Reciprocal feature learning via explicit and implicit tasks in scene text
  recognition,'' in \emph{International Conference on Document Analysis and
  Recognition}.\hskip 1em plus 0.5em minus 0.4em\relax Springer, 2021, pp.
  287--303.

\bibitem{p_spin}
C.~Zhang, Y.~Xu, Z.~Cheng, S.~Pu, Y.~Niu, F.~Wu, and F.~Zou, ``Spin:
  Structure-preserving inner offset network for scene text recognition,'' in
  \emph{Proceedings of the AAAI Conference on Artificial Intelligence}, 2021.

\bibitem{p_tps}
W.~Wang, ``Tpsnet: Thin-plate-spline representation for arbitrary shape scene
  text detection,'' \emph{arXiv preprint arXiv:2110.12826}, 2021.

\bibitem{p_seglink}
B.~Shi, X.~Bai, and S.~Belongie, ``Detecting oriented text in natural images by
  linking segments,'' in \emph{Proceedings of the IEEE conference on computer
  vision and pattern recognition}, 2017, pp. 2550--2558.

\bibitem{p_boundary}
H.~Wang, P.~Lu, H.~Zhang, M.~Yang, X.~Bai, Y.~Xu, M.~He, Y.~Wang, and W.~Liu,
  ``All you need is boundary: Toward arbitrary-shaped text spotting,'' in
  \emph{Proceedings of the AAAI conference on artificial intelligence}, 2020.

\bibitem{p_autoe}
Y.~Wang, H.~Yao, and S.~Zhao, ``Auto-encoder based dimensionality reduction,''
  \emph{Neurocomputing}, vol. 184, pp. 232--242, 2016.

\bibitem{p_multil}
D.~Ganda and R.~Buch, ``A survey on multi label classification,'' \emph{Recent
  Trends in Programming Languages}, vol.~5, no.~1, pp. 19--23, 2018.

\bibitem{d_sjsynth}
A.~Gupta, A.~Vedaldi, and A.~Zisserman, ``Synthetic data for text localisation
  in natural images,'' in \emph{GitHub}, 2016.

\bibitem{d_mjsynth}
M.~Jaderberg, K.~Simonyan, A.~Vedaldi, and A.~Zisserman, ``Synthetic data and
  artificial neural networks for natural scene text recognition,'' in
  \emph{GitHub}, 2014.

\bibitem{d_coco}
T.-Y. Lin, M.~Maire, S.~Belongie, J.~Hays, P.~Perona, D.~Ramanan,
  P.~Doll{\'a}r, and C.~L. Zitnick, ``Microsoft coco: Common objects in
  context,'' in \emph{European conference on computer vision}.\hskip 1em plus
  0.5em minus 0.4em\relax Springer, 2014, pp. 740--755.

\bibitem{p_hwrecog}
A.~Graves, M.~Liwicki, S.~Fernandez, R.~Bertolami, H.~Bunke, and
  J.~Schmidhuber, ``A novel connectionist system for unconstrained handwriting
  recognition,'' \emph{IEEE Transactions on Pattern Analysis and Machine
  Intelligence}, vol.~31, no.~5, p. 855–868, May 2009.

\bibitem{p_hogstr}
B.~Su and S.~Lu, ``Accurate scene text recognition based on recurrent neural
  network,'' \emph{Computer Vision – ACCV 2014}, p. 35–48, 2015.

\bibitem{p_crnn}
B.~Shi, X.~Bai, and C.~Yao, ``An end-to-end trainable neural network for
  image-based sequence recognition and its application to scene text
  recognition,'' \emph{IEEE transactions on pattern analysis and machine
  intelligence}, vol.~39, no.~11, pp. 2298--2304, 2016.

\bibitem{p_ctc}
\BIBentryALTinterwordspacing
A.~Graves, S.~Fern\'{a}ndez, F.~Gomez, and J.~Schmidhuber, ``Connectionist
  temporal classification: Labelling unsegmented sequence data with recurrent
  neural networks,'' in \emph{Proceedings of the 23rd International Conference
  on Machine Learning}, ser. ICML '06.\hskip 1em plus 0.5em minus 0.4em\relax
  Association for Computing Machinery, 2006, p. 369–376. [Online]. Available:
  \url{https://doi.org/10.1145/1143844.1143891}
\BIBentrySTDinterwordspacing

\bibitem{p_ace}
Z.~Xie, Y.~Huang, Y.~Zhu, L.~Jin, Y.~Liu, and L.~Xie, ``Aggregation
  cross-entropy for sequence recognition,'' in \emph{2019 IEEE/CVF Conference
  on Computer Vision and Pattern Recognition (CVPR)}, 2019, pp. 6531--6540.

\bibitem{p_tpami}
J.~Baek, G.~Kim, J.~Lee, S.~Park, D.~Han, S.~Yun, S.~J. Oh, and H.~Lee, ``What
  is wrong with scene text recognition model comparisons? dataset and model
  analysis,'' in \emph{2019 IEEE/CVF International Conference on Computer
  Vision (ICCV)}, 2019, pp. 4714--4722.

\bibitem{p_tperceptron}
\BIBentryALTinterwordspacing
L.~Qiao, S.~Tang, Z.~Cheng, Y.~Xu, Y.~Niu, S.~Pu, and F.~Wu, ``Text perceptron:
  Towards end-to-end arbitrary-shaped text spotting,'' \emph{CoRR}, vol.
  abs/2002.06820, 2020. [Online]. Available:
  \url{https://arxiv.org/abs/2002.06820}
\BIBentrySTDinterwordspacing

\bibitem{p_abcnet}
Y.~Liu, H.~Chen, C.~Shen, T.~He, L.~Jin, and L.~Wang, ``Abcnet: Real-time scene
  text spotting with adaptive bezier-curve network,'' in \emph{2020 IEEE/CVF
  Conference on Computer Vision and Pattern Recognition (CVPR)}, 2020, pp.
  9806--9815.

\bibitem{p_frcnn}
S.~Ren, K.~He, R.~Girshick, and J.~Sun, ``Faster r-cnn: Towards real-time
  object detection with region proposal networks,'' \emph{Advances in neural
  information processing systems}, vol.~28, 2015.

\bibitem{p_tcn1}
C.~Lea, M.~D. Flynn, R.~Vidal, A.~Reiter, and G.~D. Hager, ``Temporal
  convolutional networks for action segmentation and detection,'' in
  \emph{Proceedings of the IEEE Conference on Computer Vision and Pattern
  Recognition (CVPR)}, July 2017.

\bibitem{p_tcn2}
S.~Bai, J.~Z. Kolter, and V.~Koltun, ``An empirical evaluation of generic
  convolutional and recurrent networks for sequence modeling,'' \emph{arXiv
  preprint arXiv:1803.01271}, 2018.

\bibitem{p_frec}
I.~Adjabi, A.~Ouahabi, A.~Benzaoui, and A.~Taleb-Ahmed, ``Past, present, and
  future of face recognition: A review,'' \emph{Electronics}, vol.~9, no.~8, p.
  1188, 2020.

\bibitem{p_bclass}
A.~C. Ferreira, L.~R. Silva, F.~Renna, H.~B. Brandl, J.~P. Renoult, D.~R.
  Farine, R.~Covas, and C.~Doutrelant, ``Deep learning-based methods for
  individual recognition in small birds,'' \emph{Methods in Ecology and
  Evolution}, vol.~11, no.~9, pp. 1072--1085, 2020.

\bibitem{p_bilstm}
G.~Liu and J.~Guo, ``Bidirectional lstm with attention mechanism and
  convolutional layer for text classification,'' \emph{Neurocomputing}, vol.
  337, pp. 325--338, 2019.

\bibitem{p_1dconv}
S.~Kiranyaz, O.~Avci, O.~Abdeljaber, T.~Ince, M.~Gabbouj, and D.~J. Inman, ``1d
  convolutional neural networks and applications: A survey,'' \emph{Mechanical
  systems and signal processing}, vol. 151, p. 107398, 2021.

\bibitem{p_softmax}
C.~Nwankpa, W.~Ijomah, A.~Gachagan, and S.~Marshall, ``Activation functions:
  Comparison of trends in practice and research for deep learning,''
  \emph{arXiv preprint arXiv:1811.03378}, 2018.

\bibitem{p_1dattn}
H.~Xu, J.~Yang, J.~Cai, J.~Zhang, and X.~Tong, ``High-resolution optical flow
  from 1d attention and correlation,'' in \emph{Proceedings of the IEEE/CVF
  International Conference on Computer Vision (ICCV)}, October 2021, pp.
  10\,498--10\,507.

\bibitem{p_adam}
S.~Sun, Z.~Cao, H.~Zhu, and J.~Zhao, ``A survey of optimization methods from a
  machine learning perspective,'' \emph{IEEE transactions on cybernetics},
  vol.~50, no.~8, pp. 3668--3681, 2019.

\bibitem{p_bsn}
S.~Ioffe and C.~Szegedy, ``Batch normalization: Accelerating deep network
  training by reducing internal covariate shift,'' in \emph{International
  conference on machine learning}.\hskip 1em plus 0.5em minus 0.4em\relax PMLR,
  2015, pp. 448--456.

\bibitem{p_dropout}
S.~Salman and X.~Liu, ``Overfitting mechanism and avoidance in deep neural
  networks,'' \emph{arXiv preprint arXiv:1901.06566}, 2019.

\bibitem{p_sdepth}
G.~Huang, Y.~Sun, Z.~Liu, D.~Sedra, and K.~Q. Weinberger, ``Deep networks with
  stochastic depth,'' in \emph{European conference on computer vision}.\hskip
  1em plus 0.5em minus 0.4em\relax Springer, 2016, pp. 646--661.

\bibitem{p_sam}
P.~Foret, A.~Kleiner, H.~Mobahi, and B.~Neyshabur, ``Sharpness-aware
  minimization for efficiently improving generalization,'' \emph{arXiv preprint
  arXiv:2010.01412}, 2020.

\bibitem{d_text_rec_gen}
\BIBentryALTinterwordspacing
Belval, ``Textrecognitiondatagenerator: A synthetic data generator for text
  recognition,'' in \emph{GitHub}, 2021, (Accessed on 09/06/2022). [Online].
  Available: \url{https://github.com/Belval/TextRecognitionDataGenerator}
\BIBentrySTDinterwordspacing

\bibitem{p_geodl}
M.~M. Bronstein, J.~Bruna, T.~Cohen, and P.~Veli{\v{c}}kovi{\'c}, ``Geometric
  deep learning: Grids, groups, graphs, geodesics, and gauges,'' \emph{arXiv
  preprint arXiv:2104.13478}, 2021.

\bibitem{p_realtime}
M.~G. Nayagam and K.~Ramar, ``A survey on real time object detection and
  tracking algorithms,'' \emph{Int. J. Appl. Eng. Res}, vol.~10, no.~9, pp.
  8290--8297, 2015.

\bibitem{d_stdOCR}
\BIBentryALTinterwordspacing
A.~Jaiswal, ``Standard ocr dataset,'' \emph{GitHub}, 2016. [Online]. Available:
  \url{https://www.kaggle.com/preatcher/standard-ocr-dataset}
\BIBentrySTDinterwordspacing

\bibitem{d_mnist}
A.~Baldominos, Y.~Saez, and P.~Isasi, ``A survey of handwritten character
  recognition with mnist and emnist,'' \emph{Applied Sciences}, vol.~9, no.~15,
  p. 3169, 2019.

\bibitem{p_mrpune}
P.~Zhao, W.~Niu, G.~Yuan, Y.~Cai, B.~Ren, Y.~Wang, and X.~Lin, ``Achieving
  real-time object detection on mobiledevices with neural pruning search,''
  \emph{arXiv preprint arXiv:2106.14943}, 2021.

\bibitem{d_cedar}
S.~N. Srihari, S.-H. Cha, H.~Arora, and S.~Lee, ``Individuality of
  handwriting,'' \emph{Journal of forensic sciences}, vol.~47, no.~4, pp.
  856--872, 2002.

\bibitem{p_metl}
E.~Bursztein, J.~Long, S.~Lin, O.~Vallis, and F.~Chollet, ``Tensorflow
  similarity: A usable, high-performance metric learning library,''
  \emph{Fixme}, 2021.

\bibitem{p_fgsm}
I.~J. Goodfellow, J.~Shlens, and C.~Szegedy, ``Explaining and harnessing
  adversarial examples,'' \emph{arXiv preprint arXiv:1412.6572}, 2014.

\end{thebibliography}

\end{document}